\documentclass[3p]{elsarticle} 
\usepackage[hyphens]{url}

\makeatletter 
\def\ps@pprintTitle{ 
\let\@oddhead\@empty 
\let\@evenhead\@empty 
\def\@oddfoot{\hfill\thepage} 
\def\@evenfoot{\thepage\hfill}}
\makeatother

\usepackage{lineno} 
\providecommand{\tightlist}{%
  \setlength{\itemsep}{0pt}\setlength{\parskip}{0pt}}

\biboptions{sort&compress} 
\usepackage{graphicx}
\usepackage{booktabs} 

\usepackage[T1]{fontenc}
\usepackage{lmodern}
\usepackage{amssymb,amsmath}
\usepackage{ifxetex,ifluatex}
\usepackage{fixltx2e} 
\IfFileExists{upquote.sty}{\usepackage{upquote}}{}
\ifnum 0\ifxetex 1\fi\ifluatex 1\fi=0 
  \usepackage[utf8]{inputenc}
\else 
  \usepackage{fontspec}
  \ifxetex
    \usepackage{xltxtra,xunicode}
  \fi
  \defaultfontfeatures{Mapping=tex-text,Scale=MatchLowercase}
  
\fi
\IfFileExists{microtype.sty}{\usepackage{microtype}}{}
\usepackage{graphicx}
\makeatletter
\def\maxwidth{\ifdim\Gin@nat@width>\linewidth\linewidth
\else\Gin@nat@width\fi}
\makeatother
\let\Oldincludegraphics\includegraphics
\renewcommand{\includegraphics}[1]{\Oldincludegraphics[width=\maxwidth]{#1}}
\ifxetex
  \usepackage[setpagesize=false, 
              unicode=false, 
              xetex]{hyperref}
\else
  \usepackage[unicode=true]{hyperref}
\fi
\hypersetup{breaklinks=true,
            bookmarks=true,
            pdfauthor={},
            pdftitle={Reinforcement Evolutionary Learning Method for self-learning},
            colorlinks=true,
            urlcolor=blue,
            linkcolor=magenta,
            pdfborder={0 0 0}}
\begin{document}
\begin{frontmatter}

  \title{Reinforcement Evolutionary Learning Method for self-learning}
    \author[Researchers]{Kumarjit Pathak}
   \ead{Kumarjit.pathak@outlook.com} 
  
    \author[Researchers]{Jitin Kapila}
   \ead{Jitin.kapila@outlook.com} 
  
      \address[Researchers]{Data Science Researchers, Bangalore, India}
  
  \begin{abstract}
  In statistical modelling the biggest threat is concept drift which makes
  the model gradually showing deteriorating performance over time. There
  are state of the art methodologies to detect the impact of concept
  drift, however general strategy considered to overcome the issue in
  performance is to rebuild or re-calibrate the model periodically as the
  variable patterns for the model changes significantly due to market
  change or consumer behavior change etc. Quantitative research is the
  most widely spread application of data science in Marketing or financial
  domain where applicability of state of the art reinforcement learning
  for auto-learning is less explored paradigm. Reinforcement learning is
  heavily dependent on having a simulated environment which is majorly
  available for gaming or online systems, to learn from the live feedback.
  However, there are some research happened on the area of online
  advertisement, pricing etc where due to the nature of the online
  learning environment scope of reinforcement learning is explored.
  
  Our proposed solution is a reinforcement learning based, true
  self-learning algorithm which can adapt to the data change or concept
  drift and auto learn and self-calibrate for the new patterns of the data
  solving the problem of concept drift.
  \end{abstract}
  
 \end{frontmatter}


\hypertarget{introduction}{%
\section{1. Introduction}\label{introduction}}

\quad

Concept drift is well known challenge for sustainability of any machine
learning predictive model over time. Machine learning offers diverse
techniques to understand the underlying pattern of the data and
associate the same with prediction objective. Any predictive modelling
activity in either Marketing, Finance, Management are heavily dependent
on the assumption that the training data represents the pattern of
target population under specific study such as Fraud Identification,
Customer churn prediction, Marketing mix modelling, Target customer
identification for specific type of promotion etc. However due to social
\& economic development, customer behavior changes combined with other
external factors making past learned pattern, irrelevant for current
predictions. Model maintenance is one of the key activity for the
companies, wo are using machine learning for decision making. Model
maintenance involves identifying sign of concept drift and performance
decay of model periodically and re-calibrate model parameters or many a
times model complexity in case of drastic change in the data to handle
performance issue of the model. With current development of
reinforcement learning providing hope for general purpose artificial
intelligence. However, the application areas remain confined to online
learning only, where the reinforcement learning agent can learn from a
simulated environment or live environment in parallel with a stable
statistical model and learn by failing millions of times to learn about
the environment. In parallel it learns effective weight for it's own
neural network brain to represent the understanding of the environment
to be able to start performing certain activities in live environment.
Understandably so as the cost of live failure is very high, and risk
attached with the same has high penalty. Just to reinforce the point of
view, we can think of an example that if self-driving cars are trained
on live environment it would pose being threat to safety on the roads,
as well as damage of the car itself pose cost challenge.

Another example if a robot is being trained on how to jump or do some
critical work it is imperative that it is done in a simulated
environment or else cost of damage to the robot or surrounding
environment would make such experiment un-sponsorable. There are lots of
recent development happened on usage of reinforcement learning in
creating general purpose AI. Simulated environment is one of the
fundamental ask for any RL-Agent to learn. David Silver in his paper,
Silver, Schrittwieser, et al. (2017), has indicated offline learning by
observing human play video for the world champion beating algorithm
AlphaGo.

Our proposed solution is inspired from different research work on
reinforcement learning, genetic algorithm, incremental learning and
concept drift. Our goal is to create a framework which would self-learn
based on given target, self-optimize and self-calibrate in case of
performance issue due to concept drift. We used concepts from
reinforcement learning, AlphaGo, genetic algorithm along with supervised
learning to achieve our goal.

With our approach of Reinforcement Evolutionary Learning Method (RELM),
we were able to demonstrate that combinations of reinforcement learning
and genetic algorithm along with deep neural network (a.k.a matrix
function approximates), can adapt to concept drift and learn new
behavior of the data. Our main contributions are :

\begin{itemize}
\tightlist
\item
  Providing variant of genetic algorithm as an alternative to tune
  weights of the deep neural network against current successful
  back-propagation method.
\item
  Demonstration of reinforcement learning strategy for regular marketing
  analytics modelling.
\item
  Demonstration of self-learning capability to effectively handle
  concept drift issue.
\end{itemize}

\hypertarget{literature-review}{%
\section{2. Literature Review}\label{literature-review}}

\quad

Self-learning, Incremental learning, concept drift are the areas of
research since decades. \emph{Concept drift} (Gepperth and Hammer
(2016)) refers to the change in either the input data distribution or
change in the relationship between predictors and dependent variable.
Generally, two types of concept drifts observed (Virtual concept drift
\& Real concept drift). \emph{Virtual concept drift} refers to the
phenomenon that the distribution of the predictors changes over time.
\emph{Real concept drift} refers to the phenomenon that the relationship
between predictor and dependent variable p(Y\textbar{}X) itself changes.

In general concept drift gives raise to the performance issue of the
model. There are different methods to detect concept drift, one of the
popular method is to check \emph{population stability index (PSI)}. This
method (Gadidov and McBurnett 2015) checks for the performance corrosion
over consecutive time frame. We used PSI to detect real concept drift.

\quad

\[PSI = \sum_{1}^{d} (actual\ \%\ event\ \-\ expected\ \%\ event) * log \frac{actual\ \%\ event}{expected\ \%\ even}\]
\quad \qquad

where \(d\) refers to range of interval group for each feature.

Another way to identify concept drift is using statistical tests like
Hoeffding bound, by Blanco et al. (2015), or using Hellinger distance
etc. However, these are not part of the proposed RELM architecture.

Incremental learning generally refers to a continuous adaptation of the
machine learning model based on the input data changes. Gepperth and
Hammer (2016) has described an area of challenge in Incremental
learning, that is \emph{Stability-Plasticity dilemma}, which refers to
the decision dilemma, by M. Mermillod and Bonin (2013), of how much the
new concept should be learned and how much the old concept needs to be
forgotten. Experience shows that rapid adaptation gives rise to
catastrophic forgetting problem, by French (1992) \& McCloskey and Cohen
(1989). \emph{Catastrophic forgetting} refers to the fact that algorithm
might be able to learn the new concepts quickly but equally rapidly the
old information is forgotten.

Our approach is inspired by AlphaGo paper, by Silver, Schrittwieser, et
al. (2017) \&~Silver, Hubert, et al. (2017), is based on dynamic sample
generation method for experience replay in order to calibrate the old
and new learning together

Fearnet presents another brain inspired architecture proposed by Kemker
and Kanan (2018). Author has efficiently evaluated past research work
done by R. French who recommended a strategy to mitigate catastrophic
forgetting with dual separate memory centers. One with short term and
one with long term memory. Fearnet is inspired based on the rehearsal
method was proposed to have to have a mix of both new and old data to
reduce catastrophic forgetting. Generative model was proposed as an
alternative to gain memory efficiency where the algorithm can generate
random vectors of learning data from the past and augment with new data.
Fearnet was proposed inspired from brain consolidation. Fearnet uses to
complementary memory centers, HC which stores more recent information is
a probabilistic neural network. mPFC is described as old memory storage
which is a dual purpose DNN (deep Neural Network) to predict the class
with respect to the input vector and reconstruct the input using
symmetric encoder decoder. However, for marketing analytics and consumer
behavior study keeping very old information does not add value to the
predictive model as consumer preference is dynamically evolving and
hence very old data patterns becomes obsolete, however our approach is
to balance both old and new learning with incremental shift in the
sampling reference frame with respect to time.

Reinforcement learning, by Sutton and Barto (1998) \&~Silver (2015), is
an active area of research and currently acting as the backbone of the
modern artificial intelligence. Sutton explain in his book (Sutton and
Barto (1998)), different type of reinforcement learning method which can
learn without supervision with approximating the environment behavior
incrementally well, as more and more experiments happens between agent
and environment. David Silver and his team has reinforced the
applicability of reinforcement learning in complicated environments,
such as learning how to play, the game GO against the best player in the
world using reinforcement learning. Reinforcement learning by nature
adapts to the environment and create more and more accurate
representation of the environment as it interacts more and more with the
environment gives rise to the potential different application that even
if the environment behavior is changing then agent would continuously
learn to adapt to the new patterns as well, this can be used as one of
the most suitable alternative for incremental and continuous learning
which can focus on specific task. We aim to use this feature of RL to
create our framework.

David Silver in his paper (Silver, Schrittwieser, et al. (2017)) , has
showcased that with sufficient training computer can learn to play one
of the most complicated, intuitive and high variance game such as GO. He
has showcased usage of DNN in both AlphaGO and AlphaZero, Silver,
Hubert, et al. (2017) \&~Silver, Schrittwieser, et al. (2017), to learn
the game paly using raw pixcel as an input and understanding next
probable move along with the potential value of the same. Work of a
genius where he used trimmed MCTS to represent the learning by the
network and helped to have quick decision during self-play phase.
Reinforcement learning is in use for marketing analytics challenges like
dynamic pricing (Roberto Maestre 2018). Author has showcased usage of a
variant of reinforcement learning called Q-Learning with Neural Network
approximation of the value function to adaptively select the right
price. Jain index is also used to measure fairness to the pricing and
used a composite value function to learn.

Usage of reinforcement learning is becoming state of the art in
promotion strategy as well. Usage of multi arm bandit, a classical
reinforcement learning is used to personalize the promotions based on
each customer's behavior to get maximum clicks. However, the basic
assumption for applying any reinforcement learning that we are dealing
with non-IID data. In simple terms data has some sequence which is
represented as states and algorithm tries to learn the transition of
states maximizing value function. However most of the marketing research
problems deal with cross-sectional data and hence many a case the online
feedback is not possible, and it becomes a bottle-neck for applicability
of reinforcement learning. Our approach postulates the problem of
classification modelling in any marketing analytics/financial analysis
scenario as a policy gradient RL method.

We also explored the current application of genetic algorithm on the
field of deep learning and artificial intelligence as heuristic search
method. Till today back propagation has undoubtedly achieved high
success and all latest and greatest algorithms are learning based on
back propagation. Paul Werbos propose that backpropagation could be used
for neural nets after analyzing it in depth in his 1974 PhD Thesis. In
1986 through the work of David E. Rumelhart, Geoffrey E. Hinton, Ronald
J. Williams, and James McClelland, backpropagation gained recognition.
Currently all the state of the art methods (Vu N.P. Dao 2001) in deep
learning majorly trains the parameters using backpropagation method.
However, there are active research happening to overcome the limitations
of backpropagation in different ways. Limitations are majorly vanishing
gradient problem, which occurs due to high number of layers used in deep
learning architecture. Gradient updates may get low value while back
propagation in the initial layers compared to the layers near to the
output. To mitigate this problem, skip connections are suggested in
FRCNN papers also concepts like LSTM came up. Another problem area which
sometimes becomes critical, is to have a large unbiased dataset to train
a large neural network. Neural network has tendency of overfitting if
the data is not sufficiently large and success of the generalization
aspect of back propagations is majorly controlled by the size and the
quality of the input data.

We explored all recent applications of reinforcement learning where the
value function approximation or the policy function approximation is
done using deep learning architecture. This has made the algorithm so
powerful that with mare visual input like videos and pictures the
algorithm is able to decide the actions suitable of current state
keeping in mind long term rewards of winning the game. After doing
considerable amount of study we could see that the function
approximations are driven by backpropagation method in all recent
reinforcement learning application. Our approach here proposes an
architecture to have the function approximator learn through a variant
of genetic algorithm which gives the algorithm more capability to learn
and explore and adapt to the changed scenario of market. Related work
was done by Felipe Petroski Such (Felipe Petroski Such 2018) to showcase
applicability of genetic algorithm augmented learning instead of back
propagation with distributed computing using either multiple CPU or GPU.
Author has showcased a novel method for distributed deep genetic
algorithm where large parameter vectors are represented by a function
with initial seed and list of random seeds which produces the series of
mutation. This is done using a deterministic mutation function.

We have used CMA-ES
\(covariance\ matrix\ adaptation\ evolution\ strategy\) for the learning
of algorithm. Genetic algorithm augmentation on deep learning has been
gaining momentum recently and many researches are coming up in different
development. Genetic Algorithm is majorly used in two ways, evolving the
architecture/topology of the neural network \& weight updates of the
network. Application of genetic algorithm in neural network for
optimization has proven to be great. Pattern recall analysis by Kumar
and Pratap (2010), showcases uses of GA to update the optimal weights.
Stoc price index prediction using GA augmented neural network achieved
better result. Cervical cancer detection by P. Mitra (2001), also shows
close to 90\% accuracy. There are number of applications showcased by
different researchers are given in Appendix-1 where success of genetic
algorithm augmented neural network has proven to have better accuracy
than training the network with back propagation.

\hypertarget{method}{%
\section{3. Method}\label{method}}

\quad

Our solution framework works on the concept of reinforcement learning
with policy gradient. Idea being formulating the regular quantitative
marketing and financial analysis problems with cross sectional data as a
deep evolutionary reinforcement learning problem to overcome the issue
of \emph{``Concept Drift''}. Challenge of self-learning, auto-learning
and concept drift is wide across industries. Our solution has the
capability to avoid model maintenances activity and have the model
automatically tune itself for any sort of classification model using
quantitative data.

\hypertarget{refinforcement-learning}{%
\subsubsection{Refinforcement Learning
:}\label{refinforcement-learning}}

\quad

Reinforcement learning refers to an active learning(RL) method without
supervision based on either a value function or policy function. RL
Agent experiments to understand the environment and store the learning
based on the rewards continuously and observe the state transition. This
helps agent to make an approximately representative model of the
environment behavior. This means the environment becomes predictable for
the agent and now at any given state agent would know what the best
possible action is to take to get optimum long-term reward.
Reinforcement learning exploits the Markov property and with
approximation function tries to achieve the same. Markov property refers
to the fact the current state information is sufficient to predict the
best action and the next state transition. Reinforcement learning uses
Bellman equation heavily to iteratively update the representation or
approximation of the environment using policy approximator or value
approximator or both. \quad

\[Value_{State\ S}^{Policy\ \pi} = Reward_{\ state\ =\ s}^{action\ =\ a} +\delta \sum_{s'\epsilon  S} P_{ss'} V(s')\]

where \(s\) is the current state \(s'\) is the next state, \(P_{ss’}\) =
probability of the state transition and Policy = \(\pi\)

These approximation function or model is currently done using deep
learning models which helps to have the universal function approximation
capability and based on the gradient update during back propagation it
updates the representation.

This policy function and the value function approximations are now done
using deep neural networks giving rise to the concept of Deep
Reinforcement Learning.

Different architecture and methodology has been researched over decades
such as Q- learning, Deep Q learning, Dueling Deep Q learning, Monte
Carlo learning, Temporal Differencing Learning and SARSA, etc. to learn
based on the state or action value function. However, in many cases
understanding the state value becomes difficult due to continuous states
or very high number of states. This gives rise to policy gradient
methods.

\hypertarget{refinforcement-learning-1}{%
\subsubsection{Refinforcement Learning
:}\label{refinforcement-learning-1}}

\quad

The method is based in directly parameterizing the policy. This means
what action to take being on the current state is a direct output from
the function approximator itself. Hence, the concepts tends to get rid
of the limitation of value based learning. Value based learning has a
limitation of near deterministic policy either \(greedy\) or
\(\epsilon\ - greedy\) .It is observed that due to the imbalance of
exploration vs exploitation, in many cases the value based approximation
get's stuck in the local optimum.

Goal of policy gradient method is to find best parameters to optimize
policy at each state and we can measure the quality of the policy using
:

\begin{itemize}
\tightlist
\item
  Start value in case of episodic environment
\item
  Average value for continuous environment
\item
  Average reward per time step
\end{itemize}

Advantage of this method has been really promising.

\begin{itemize}
\tightlist
\item
  It generally finds better convergence.
\item
  Very effective in case if the state space is high dimensional or
  continuous
\item
  Main advantage is that it can learn stochastic policies. Policy can
  change due to any factor , however this advantage is not there in
  value based reinforcement learning method.
\end{itemize}

It suffers from few disadvantages as well:

\begin{itemize}
\tightlist
\item
  Policy evaluation is high variance and becomes inefficient with
  respect to state sometimes.
\item
  Local optimum issue
\end{itemize}

Different methods under policy gradient been researched over time such
as \textbf{``Deep Policies''}, which uses a neural network to represent
a policy output model. \(Action = f(state, weight)\). Action is a
function of state and parameter weight. Objective is to optimize total
discounted reward.

\textbf{DDPG (Deterministic Deep Policy Gradient)} method uses an actor
and critic network where actor is a policy network and critic being
q-learning based network. This method uses ``Experience Replay'' for
both actor and critic with a target frozen for an episode of learning to
avoid dynamic oscillation of the target each time of update and avoid
oscillation of target. As oscillation of target at each iteration makes
the learning instable due to frequent dynamic variation.

\hypertarget{procedure}{%
\section{4. Procedure}\label{procedure}}

\hypertarget{rlem-architecture}{%
\subsubsection{\texorpdfstring{\textbf{RLEM-architecture}:}{RLEM-architecture:}}\label{rlem-architecture}}

\quad

Our innovative approach formulates a different variant of ``Deep
Policy'' agent augmented with experience replay and CMA-ES (Covariance
Matrix Adaptation Evolution Strategy) for the policy function
optimization. Using experience replay gives us the potential to learn
from new stream of data on continuous basis. Instead of gradient
backpropagation-based learning we have used CMA-ES to mitigate the
problem with policy gradient method being stuck in local optimum. This
also increases the efficiency of the learning and faster convergence.

CMA-ES takes the result of each generation and adaptively increase and
decrease the search space for next generation. It calculates and adapt
for the mean variance and calculate the entire covariance matrix for
update. Thus, it understands the relations of intermediate generations'
genomes. CMA-ES modifies the covariance calculation in intelligent way
to help the optimization. This algorithm first focuses on top N-best
solutions (genomes) from the current generation it calculates the
covariance matrix for next generation. Bottleneck of this optimization
method is the size of the covariance metrics. To overcome the size of
the covariance matrix which is dependent on the number of parameters we
used concept of latent space modelling to reduce the number of dimension
at the first go. This also enables us to model based on the latent
feature and reduce the noise in the data variations which is evident for
any marketing research problem.

\begin{figure}
\centering
\includegraphics{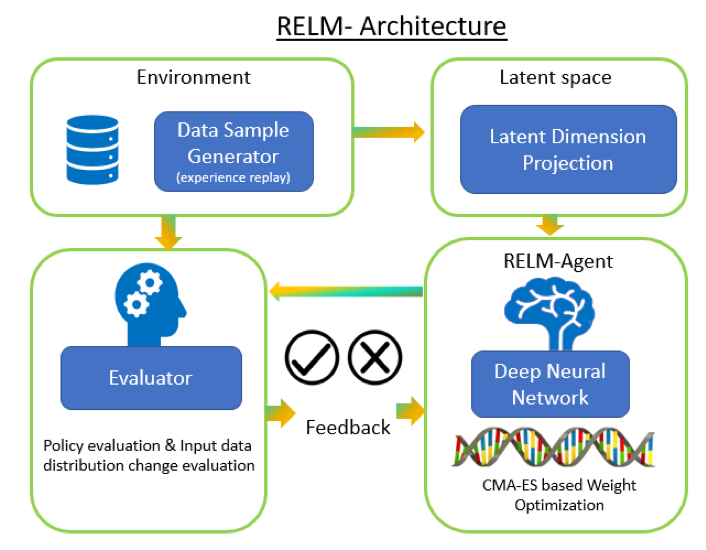}
\caption{Architecture of RELM}
\end{figure}

\quad

\hypertarget{rlem-environment}{%
\subsubsection{\texorpdfstring{\textbf{RLEM-Environment}:}{RLEM-Environment:}}\label{rlem-environment}}

\quad

By nature reinforcement learning is an online learning algorithm where
agent learns from the interaction with the environment. This area was
not part of active research and hence applicability of reinforcement
learning for regular classification modelling in Marketing analytics was
not explored to the best of it's possibility. We intend to break this
paradigm and hence we created an environment with the initial static
data of any regular classification modelling. Idea being representing
the input vector of each row in the data-frame as \(state(S)\) and the
right output label as the right \(action(a)\).

Data sample generator is used to generate sample for current learned
data points for the algorithm so that the old learning is not forgotten
by the network. We don't intend to keep all the data points from every
time period and hence created an window shift mechanism for the data.
Data with more than \(t\) period old are discarded automatically from
the environment. Time \(t\) is adjustable. This is just to avoid memory
issues and performance issues keeping in mind that for marketing
analysis data beyond a certain period does not relevant for learning due
to continuous shift in socio-economic development and change in customer
behavior.

\hypertarget{laten-space}{%
\subsubsection{\texorpdfstring{\textbf{Laten
Space}:}{Laten Space:}}\label{laten-space}}

\quad

Marketing research problem generally deals with two kind of data type,
discrete and continuous. However, we used this block of the architecture
to transform both the data type into a continuous and reduced dimensions
and extract relevant information's without noise for the state
representation which is our primary assumption for using \emph{``Deep
Policy''} method of reinforcement learning. We segregated discrete and
continuous features from the data frame and separately utilized at this
layer.

For continuous data we used Variational Auto Encoder,by Kingma and
Welling (2013) \&~Pu et al. (2016), to generate the bottleneck layer of
representation eliminating noise from input feature set. For discrete
data we used Restricted Boltzmann Machine Salakhutdinov, Mnih, and
Hinton (2007) to bring the discrete variables in continuous reduced
dimensional representation.

\hypertarget{relm-agent}{%
\subsubsection{\texorpdfstring{\textbf{RELM
Agent}:}{RELM Agent:}}\label{relm-agent}}

\quad

his block represents a deep neural network with layers of tanh
activation and sigmoid activation consisting of
INPUT-FC{[}45{]}-tanh-FC{[}15{]}-tanh-FC{[}6{]}-tanh-FC{[}1{]}-sigmoid-OUTPUT.
This deep learning model takes input from the latent representation of
each vector (`S' = latent state representation) and produces action (`a'
= Y = output feature). Learning mechanism is augmented with CMA-ES,
hence it is gradient free heuristic search.

\hypertarget{evaluator}{%
\subsubsection{\texorpdfstring{\textbf{Evaluator}:}{Evaluator:}}\label{evaluator}}

\quad

This block is responsible maintaining the sanity of the environment and
supervise the overall activity of the architecture. During the initial
phase when agent is learning from self-play with the environment it
checks if the performance of the agent is increasing or constant over
time. It triggers to save the best solutions periodically wherever it
shows improvement. During the new data input which environment has not
seen earlier and does not belong to the current environment it evaluates
the input distribution of each feature and measure ``Accuracy'',
``F1-measure'' and PSI (Population Stability Index) value. It checks if
agent is showing a performance drift and imitate the agent for
re-training based on the new data and label. During any production
environment this facility can be directly used if feedback is sent back
to Evaluator on the accuracy of last batch prediction. This will create
a feedback-loop. Based on the periodic offline or online feedback,
algorithm would start tuning itself to adjust to new data patterns again
and again.\\
\quad

\begin{figure}
\centering
\includegraphics{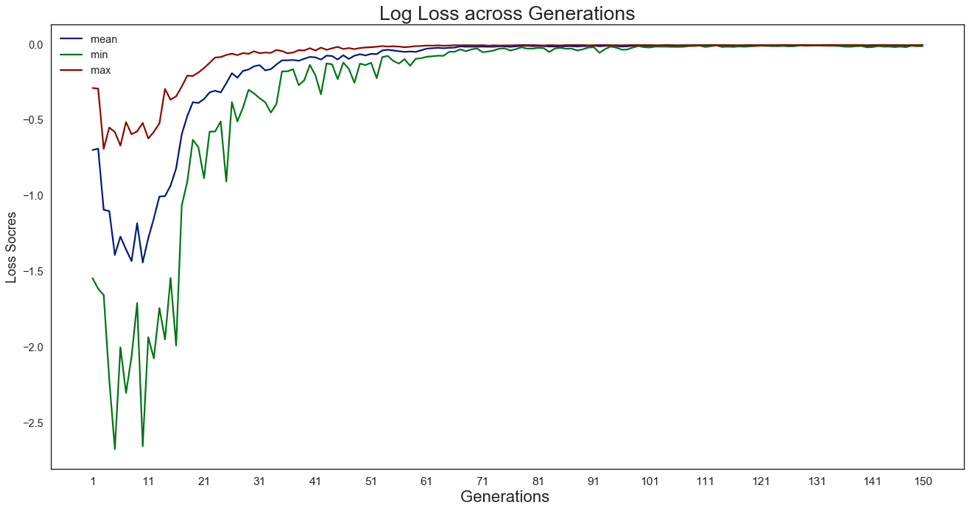}
\caption{Log Loss Across Generations}
\end{figure}

\quad

\begin{figure}
\centering
\includegraphics{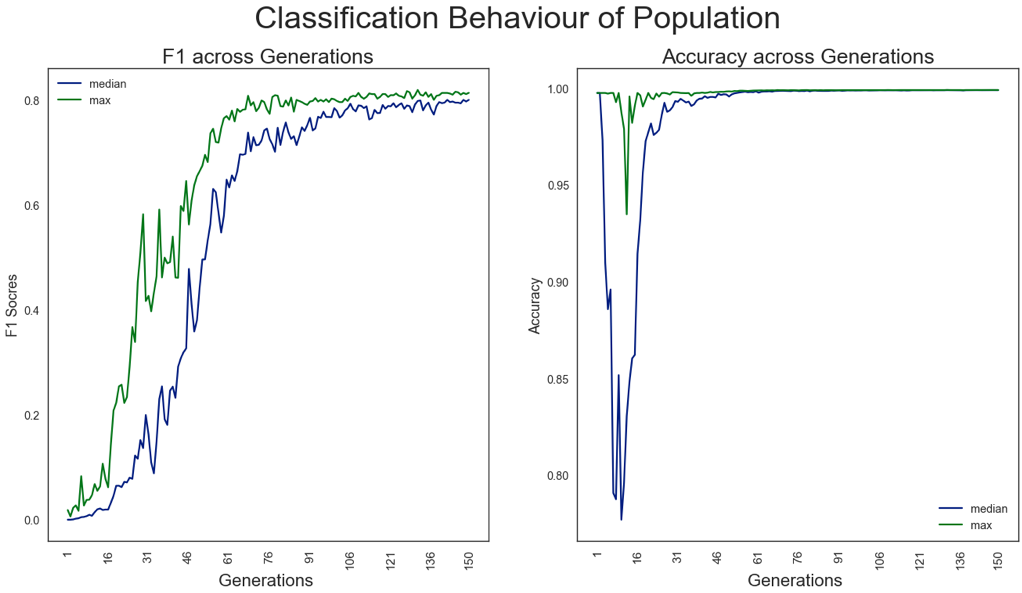}
\caption{Accuracy and F1 across Generations}
\end{figure}

\hypertarget{problem-formulation}{%
\subsubsection{\texorpdfstring{\textbf{Problem
Formulation}:}{Problem Formulation:}}\label{problem-formulation}}

\quad

Let's assume a classification modelling in either marketing or financial
domain. We try to solve any supervised classification with dependent
variable \(Y\) and independent variables \(x1,x2,x3,x4,x5,x6….\).
assuming a mathemetical non linear or linear relationship between \(Y\)
and \(X_s\). Let's assume the mathematical function \(Y=f(X, w)\) where
\(X\) represents independent feature matrix, w- represents the weight
matrix and Y represents Dependent feature matrix. Our hypothesis , we
can represent each vector of the observation with respect to the
independent variables as vector of \(\vec{X}\) . So for N observations
we have N vector of Xs and Ys. Vector of Xs can be assumed to be the
continuous states (in the latent space).\\
Now it can be safely assumed to have met the criteria for ``Deep
Policy''" learning algorithm. Deep policy learning represents a
functional equation between \(state(S=\vec{X})\) and \(weight(w)\) to
solve the \(action(a)\).

\[a = f(S,w)\]

Objective function is taken here is undiscounted reward function based
on number of right predictions and the F1 score of the prediction.

\quad

\hypertarget{experminetal-results}{%
\section{5. Experminetal Results}\label{experminetal-results}}

\quad

We tested this concept on different industrial data with consistent
results, however due to sensitivity of the data showcasing the same is
not considered. To showcase our result we using public data related to
credit card fault detection from Kaggle. This data comes with the PCA
features already and hence already in the latent space. To stop
repeating the latent dimension projection, we muted the latent space
block and went ahead to apply concept of RELM. Results were encouraging.
Fig 2 and 3 represents the initial learning of the RELM agent. As we can
see within 40th generation accuracy becomes stable. Objective function
is also comprising of F1-score and hence post 40th generation
improvement on accuracy was very less however, algorithm starts to tune
itself for F1 score keeping accuracy intact.
\quad

To prove that RELM's capability to adapt to concept drift we kept 70K
records separately not being part of the reinforcement environment
during initial training. Once the agent is comfortable with the current
data we start pushing the holdout data to RELM environment. \quad

\begin{figure}
\centering
\includegraphics{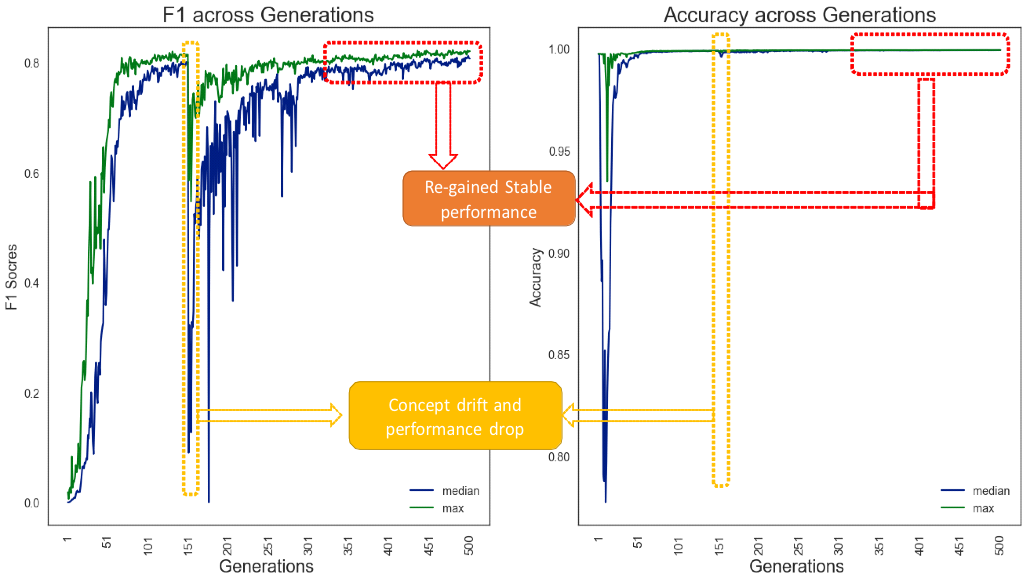}
\caption{Illus. 1 - Showcasing RELM result with concept drift entry and
how it calibrates itself by learning from new data}
\end{figure}

\quad

\begin{figure}
\centering
\includegraphics{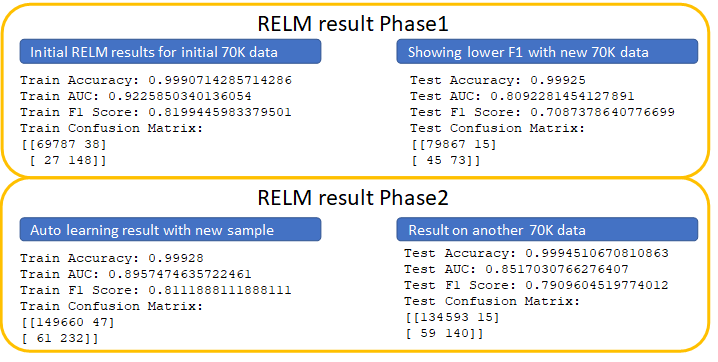} 
\caption{Illus. 2 - Showcasing RELM result with concept drift entry and
how it calibrates itself by learning from new data}
\end{figure}

Fig 4 and 5 shows that post initial training when RELM encounters the new data it had a performance decay which was caught by the evaluator and initiated the re-calibration of the RL agent. Due to the re-calibration to new data patterns algorithm
starts adapting to the new trends and regain it's confidence on the
environment. \quad

\hypertarget{conclusion-and-future-scope}{%
\section{6. Conclusion and Future
Scope}\label{conclusion-and-future-scope}}

From our study and analysis, we are confident that RELM architecture can
be used industrially for solving the issue of concept drift in any
classification modelling related marketing or financial domain with
quantitative data. We also break the paradigm of not using reinforcement
learning for offline learning. RELM can self-learn, self-optimize,
self-calibrate based the changes in the data and issue in performance.
With this auto-ML capability we open the door for automating the model
maintenance activity and provide a flexible and powerful solution each
time of customer preference and market scenario changes.

\hypertarget{appendix--1-list-of-references-where-ann-with-genetic-algorithm-has-been-used}{%
\section{Appendix -1 : List of References where ANN with genetic
Algorithm has been
used}\label{appendix--1-list-of-references-where-ann-with-genetic-algorithm-has-been-used}}

\begin{enumerate}
\def\labelenumi{\arabic{enumi})}
\tightlist
\item
  V. Bevilacqua, G. Mastronardi andF. Menolascina, Genetic Algorithm and
  Neural Network Based Classification in Microarray Data Analysis with
  Biological Validity Assessment, Berlin Heidelberg: Springer-Verlag,
  475-484 (2006).
\item
  H. Karimi and F. Yousefi, Application of artificial neural
  networkgenetic algorithm (ANNGA) to correlation of density in
  nanofluids, Fluid Phase Equilibr. 336, 79-83 (2012).
\item
  L. Jianfei, L. Weitie, C. Xiaolong andL. Jingyuan, Optimization of
  Fermentation Media for Enhancing Nitrite-oxidizing Activity by
  Artificial Neural Network Coupling Genetic Algorithm, Chinese Journal
  of Chemical Engineering, 20, 5, 950-957 (2012).
\item
  G. Kim, J. Yoona, S. Ana, H. Chob and K. Kanga, Neural network model
  incorporating a genetic algorithm in estimating construction costs.
  Build Env. 39, 1333-1340 (2004).
\item
  H. Kim, K. Shin andK. Park, Time Delay Neural Networks and Genetic
  Algorithms for Detecting Temporal Patterns in Stock Markets, Berlin
  Heidelberg: Springer-Verlag, 1247-1255 (2005).
\item
  Kim H, Shin K (2007) A hybrid approach based on neural networks and
  genetic algorithms for detecting temporal patterns in stock markets.
  Appl Soft Comput 7:569-576
\item
  K.P. Ferentinos, Biological engineering applications of feedforward
  neural networks designed and parameterized by genetic algorithms,
  Neural Network, 18, 934-950 (2005).
\item
  K. Kim and I. Han, Genetic algorithms approach to feature
  discretization in artificial neural networks for the prediction of
  stock price index, Expert Syst. Appl. 19,125-132(2000).
\item
  M.A.~Ali and S.S. Reza, Intelligent approach for prediction of minimum
  miscible pressure by evolving genetic algorithm and neural network,
  Neural Comput. Appl. DOI 10.1007/s00521-012-0984-4, 1-8 (2013).
\item
  Z.M.R. Asif, S. Ahmad and R. Samar, Neural network optimized with
  evolutionary computing technique for solving the 2-dimensional Bratu
  problem, Neural Comput. Appl. DOI 10.1007/s00521-012-1170- 4 (2012).
\item
  S. Ding, Y. Zhang, J. Chen andW. Jia, Research on using genetic
  algorithms to optimize Elman neural networks, Neural Comput.
  Appl.23,2, 293-297 (2013).
\item
  Y. Feng, W. Zhang, D. Sun andL. Zhang, Ozone concentration forecast
  method based on genetic algorithm optimized back propagation neural
  networks and support vector machine data classification, Atmos.
  Environ. 45, 1979-1985 (2011).
\item
  W. Ho andC. Chang, Genetic-algorithm-based artificial neural network
  modeling for platelet transfusion requirements on acute myeloblastic
  leukemia patients, Expert Syst. Appl. 38, 6319-6323 (2011).
\item
  G. Huse, E. Strand andJ. Giske, Implementing behavior in
  individual-based models using neural networks and genetic algorithms,
  Evol. Ecol. 13, 469- 483 (1999).
\item
  A. Johari, A.A. Javadi and G. Habibagahi, Modelling the mechanical
  behaviour of unsaturated soils using a genetic algorithm-based neural
  network, Comput. Geotechn. 38, 2-13 (2011).
\item
  R.J. Kuo, A sales forecasting system based on fuzzy neural network
  with initial weights generated by genetic algorithm, Eur. J. Oper.
  Res. 129, 496-517 (2001).
\item
  Z. Liu,A. Liu, C.Wang andZ. Niu, Evolving neural network using real
  coded genetic algorithm (GA) for multispectral image classification,
  Future Gener. Comput. Syst. 20, 1119-1129 (2004).
\item
  L. Xin-lai, L. Hu, W. Gang-lin andW.U. Zhe, Helicopter Sizing Based on
  Genetic Algorithm Optimized Neural Network, Chinese J. Aeronaut. 19,3,
  213-218 (2006).
\item
  H. Mahmoudabadi, M. Izadi and M.M. Bagher, A hybrid method for grade
  estimation using genetic algorithm and neural networks, Comput.
  Geosci. 13, 91-101 (2009).
\item
  Y. Min, W. Yun-jia andC. Yuan-ping, An incorporate genetic algorithm
  based back propagation neural network model for coal and gas outburst
  intensity prediction, ProcediaEarth. Pl. Sci. 1, 12851292 (2009).
\item
  P. Mitra, S. Mitra, S.K. Pal, Evolutionary Modular MLP with Rough Sets
  and ID3 Algorithm for Staging of Cervical Cancer, Neural Comput. Appl.
  10, 67- 76(2001).
\item
  M. Nasseri, K. Asghari and M.J. Abedini, Optimized scenario for
  rainfall forecasting using genetic algorithm coupled with artificial
  neural network, Expert Syst. Appl. 35, 1415-1421 (2008).
\item
  P.C. Pendharkar, Genetic algorithm based neural network approaches for
  predicting churnin cellular wireless network services, Expert Syst.
  Appl. 36, 6714- 6720 (2009).
\item
  H. Peng and X. Ling, Optimal design approach for the plate-fin heat
  exchangers usingneural networks cooperated with genetic algorithms,
  Appl. Therm. Eng. 28, 642-650 (2008).
\item
  S. Koer and M.C. Rahmi, Classifying Epilepsy Diseases Using Artificial
  Neural Networks and Genetic Algorithm, J. Med. Syst. 35, 489-498
  (2011).
\item
  A. Sedki, D. Ouazar and E. El Mazoudi, Evolving neural network using
  real coded genetic algorithm for daily rainfallrunoff forecasting,
  Expert Syst. Appl. 36, 4523-4527(2009)
\end{enumerate}

\hypertarget{references}{%
\section*{References}\label{references}}
\addcontentsline{toc}{section}{References}

\hypertarget{refs}{}
\leavevmode\hypertarget{ref-13drifthoeff}{}%
Blanco, Isvani Inocencio Frías, José del Campo-Ávila, Gonzalo
Ramos-Jiménez, Rafael Morales Bueno, Agustín Alejandro Ortiz Díaz, and
Yailé Caballero Mota. 2015. ``Online and Non-Parametric Drift Detection
Methods Based on Hoeffding's Bounds.'' \emph{IEEE Transactions on
Knowledge and Data Engineering} 27: 810--23.

\leavevmode\hypertarget{ref-21gaNetwork}{}%
Felipe Petroski Such, Edoardo Conti, Vashisht Madhavan. 2018. ``Deep
Neuroevolution: Genetic Algorithms Are a Competitive Alternative for
Training Deep Neural Networks for Reinforcement Learning.''
\emph{Arxiv.org}.

\leavevmode\hypertarget{ref-6catforget1}{}%
French, Robert. 1992. ``Semi-Distributed Representations and
Catastrophic Forgetting in Connectionist Networks.'' \emph{Connection
Science} 4: 1--10.

\leavevmode\hypertarget{ref-12stability}{}%
Gadidov, Bogdan, and Benjamin McBurnett. 2015. ``Population Stability
and Model Performance Metrics Replication for Business Model at Suntrust
Bank.'' \emph{SESUG} 4.

\leavevmode\hypertarget{ref-1conceptdrift}{}%
Gepperth, Alexander, and Barbara Hammer. 2016. ``Incremental Learning
Algorithms and Application.'' \emph{ESANN, Computational Intelligence
and Machine Learning}, 1--12.

\leavevmode\hypertarget{ref-11fearnet}{}%
Kemker, Ronald, and Christopher Kanan. 2018. ``FearNet: Brain-Inspired
Model for Incremental Learning.''
\url{https://openreview.net/forum?id=SJ1Xmf-Rb}.

\leavevmode\hypertarget{ref-50vae}{}%
Kingma, Diederik P, and Max Welling. 2013. ``Auto-Encoding Variational
Bayes.'' \emph{arXiv Preprint arXiv:1312.6114}.

\leavevmode\hypertarget{ref-49nngagrade}{}%
Kumar, S., and M.S. Pratap. 2010. ``Pattern Recall Analysis of the
Hopfield Neural Network with a Genetic Algorithm.'' \emph{Comput. Math.
Application} 60: 1049--57.

\leavevmode\hypertarget{ref-7catforget2}{}%
McCloskey, Michael, and Neal J. Cohen. 1989. ``Catastrophic Interference
in Connectionist Networks: The Sequential Learning Problem.'' Edited by
Gordon H. Bower, Psychology of learning and motivation, 24. Academic
Press: 109--65.
\url{https://doi.org/https://doi.org/10.1016/S0079-7421(08)60536-8}.

\leavevmode\hypertarget{ref-9plasticity}{}%
M. Mermillod, A. Bugaiska, and P. Bonin. 2013. ``The
Stability-Plasticity Dilemma: Investigating the Continuum from
Catastrophic Forgetting to Age-Limited Learning Effects.''
\emph{Frontiers in Psychology} 4: 504--6.

\leavevmode\hypertarget{ref-42nngacancer}{}%
P. Mitra, S.K. Pal, S. Mitra. 2001. ``Evolutionary Modular Mlp with
Rough Sets and Id3 Algorithm for Staging of Cervical Cancer.''
\emph{Neural Comput. Appllication} 10: 60--76.

\leavevmode\hypertarget{ref-51vaedl}{}%
Pu, Yunchen, Zhe Gan, Ricardo Henao, Xin Yuan, Chunyuan Li, Andrew
Stevens, and Lawrence Carin. 2016. ``Variational Autoencoder for Deep
Learning of Images, Labels and Captions.'' Edited by Daniel D. Lee,
Masashi Sugiyama, Ulrike von Luxburg, Isabelle Guyon, and Roman Garnett,
2352--60.
\url{http://dblp.uni-trier.de/db/conf/nips/nips2016.html\#PuGHYLSC16}.

\leavevmode\hypertarget{ref-16rldynamic}{}%
Roberto Maestre, Alberto Rubio, Juan Duque. 2018. ``Reinforcement
Learning for Fair Dynamic Pricing.'' \emph{Intelligent Systems
Conference, London, UK}, 1--7.

\leavevmode\hypertarget{ref-52rbm}{}%
Salakhutdinov, Ruslan, Andriy Mnih, and Geoffrey E. Hinton. 2007.
``Restricted Boltzmann Machines for Collaborative Filtering.'' Edited by
Zoubin Ghahramani, ACM international conference proceeding series, 227
(October). ACM: 791--98.
\url{http://dblp.uni-trier.de/db/conf/icml/icml2007.html\#SalakhutdinovMH07}.

\leavevmode\hypertarget{ref-19rltut}{}%
Silver, David. 2015. ``Reinforcement Learning.''
\url{http://www0.cs.ucl.ac.uk/staff/d.silver/web/Teaching.html}.

\leavevmode\hypertarget{ref-2david1}{}%
Silver, David, Thomas Hubert, Julian Schrittwieser, Ioannis Antonoglou,
Matthew Lai, Arthur Guez, Marc Lanctot, et al. 2017. ``Mastering Chess
and Shogi by Self-Play with a General Reinforcement Learning
Algorithm.'' \emph{CoRR} abs/1712.01815.
\url{http://dblp.uni-trier.de/db/journals/corr/corr1712.html\#abs-1712-01815}.

\leavevmode\hypertarget{ref-3david2}{}%
Silver, David, Julian Schrittwieser, Karen Simonyan, Ioannis Antonoglou,
Aja Huang, Arthur Guez, Thomas Hubert, et al. 2017. ``Mastering the Game
of Go Without Human Knowledge.'' \emph{Nature} 550 (October). Macmillan
Publishers Limited, part of Springer Nature. All rights reserved.: 354.
\url{http://dx.doi.org/10.1038/nature24270}.

\leavevmode\hypertarget{ref-18book}{}%
Sutton, Richard S., and Andrew G. Barto. 1998. \emph{Reinforcement
Learning: An Introduction}. MIT Press.
\url{http://www.cs.ualberta.ca/~sutton/book/the-book.html}.

\leavevmode\hypertarget{ref-17performbp}{}%
Vu N.P. Dao, Rao Vemuri. 2001. ``A Performance Comparison of Different
Back Propagation Neural Networks Methods in Computer Network Intrusion
Detection.'' \emph{Cyber Defense Initiative, Washington}, 3--*.

\end{document}